\documentclass[final,12pt]{clear2025} 

\title{Multi-Domain Causal Discovery in Bijective Causal Models}
\usepackage{times}

\clearauthor{%
	  \Name{Kasra Jalaldoust} \Email{kasra@cs.columbia.edu}\\
        \addr Department of Computer Science, Columbia University
	 \AND
	 \Name{Saber Salehkaleybar}
\Email{s.salehkaleybar@liacs.leidenuniv.nl}\\
     \addr Leiden Institute of Advanced Computer Science (LIACS), Leiden University 
     \AND
	 \Name{Negar Kiyavash} \Email{negar.kiyavash@epfl.ch}\\
  \addr College of Management of Technology, École Polytechnique Fédérale de Lausanne (EPFL)
}

\usepackage[utf8]{inputenc} 
\usepackage[T1]{fontenc}    
\usepackage{hyperref}       
\usepackage{url}            
\usepackage{booktabs}       
\usepackage{amsfonts}       
\usepackage{nicefrac}       
\usepackage{microtype} 
\usepackage{natbib}

\usepackage[mathscr]{euscript}

\usepackage{amsmath}
\newtheorem{thm}{Theorem}
\usepackage{amssymb}
\usepackage{algorithm}
\usepackage{algorithmic}

\newcommand\independent{\protect\mathpalette{\protect\independenT}{\perp}}
\def\independenT#1#2{\mathrel{\rlap{$#1#2$}\mkern2mu{#1#2}}}

\newtheorem{defn}{Definition}
\newtheorem{lem}{Lemma}
\newtheorem{assm}{Assumption}

\newtheorem{cor}{Corollary}
\newtheorem{prop}{Proposition}
\newtheorem{rem}{Remark}

\usepackage{siunitx}

\newcommand*{\specialcell}[2][b]{%
	\begin{tabular}[#1]{@{}c@{}}#2\end{tabular}%
}
\newcommand*{\specialcellbold}[2][b]{%
	\bfseries
	\sisetup{text-rm = \bfseries}%
	\begin{tabular}[#1]{@{}c@{}}#2\end{tabular}%
}

\begin{document}
	
	\maketitle
	\begin{abstract}
		We consider the problem of causal discovery (a.k.a., causal structure learning) in a multi-domain setting. We assume that the causal functions are invariant across the domains, while the distribution of the exogenous noise may vary. Under causal sufficiency (i.e., no confounders exist), we show that the causal diagram can be discovered under less restrictive functional assumptions compared to previous work. What enables causal discovery in this setting is bijective generation mechanisms (BGM), which ensures that the functional relation between the exogenous noise $E$ and the endogenous variable $Y$ is bijective and differentiable in both directions at every level of the cause variable $X = x$. BGM generalizes a variety of models including additive noise model, LiNGAM, post-nonlinear model, and location-scale noise model. Further, we derive a statistical test to find the parents set of the target variable. Experiments on various synthetic and real-world datasets validate our theoretical findings. 
	\end{abstract}

	\section{Introduction}\label{introduction}
	Recovering causal relations is central to many scientific fields. Causal relationships between random variables are commonly represented by a graph called causal diagram in which a directed edge from variable $X$ to variable $Y$, if $X$ is a direct cause of $Y$. Performing controlled experiments can recover the causal relationships, however, it is not always possible to perform controlled experiments due to technical or ethical considerations. This limits us to use the observational data that is collected passively from the variables in the system in order to recover the causal diagram. From the observational data, without parametric assumptions, the true causal diagram can be identified up to a Markov equivalence class (MEC), which is the class of graphs representing the same set of conditional independence relations among the observed variables \citep{PL09}. In order to identify the MEC, one can use constraint-based algorithms such IC and IC* \cite{PL09}, PC, FCI \cite{SP93} or score-based methods such as greedy equivalence search \cite{CH03}. Moreover, within the framework of structural causal models (SCM), the true graph can be recovered uniquely by considering some additional assumptions on the data generation mechanisms such as non-Gaussianity of noise \cite{SH06}, or non-linearity of functional relations \cite{HO09}. The aforementioned results rely on Markovianity assumption, which states that there exist no exogenous variables that affect multiple endogenous variables, i.e., no confounders; throughout this paper we assume Markovianity.

	In recent years, causal discovery using data collected from multiple domains has gained attention \citep{jaber2020causal,perry2022causal,huang2019causal,peters2016causal,mooij2020joint,yang2018characterizing,squires2020permutation}. This setting is commonly formulated as having i.i.d samples of the variables across different domains, such that the causal modules may change across datasets, and these changes may or may not be known to the learner. As an example from applications, in \citep{ackerer2020deep} heteroskedasticity has been observed in food expenditure as a function of income for Belgian working class households. From the methods, \cite{JP16} introduced the notion of ``invariant prediction'', where it is assumed that for every domain $e \in \mathcal{E}$, the response variable $Y^i$ has a fixed set of predictors $X_{S^*}, S^* \subset \{  1, 2, \dots, p \}$ as its ``invariant" predictors, such that the residual distribution is invariant across the domains when $X_{S^*}$ is the set of explanatory variables. The task is then finding the set of covariates $S^*$ using observations from multiple domains. Although, correctly recovering the set $S^*$ is contingent on (1) having access to large enough set of domains and (2) invariance of the mechanism that determines $Y$ based on  $X_{S^*}$ across all domains. The proposed method searches exhaustively over all possible subsets of the covariates, thus, has high computational and statistical complexity. The output is guaranteed to be contained by $X_{S^*}$ with high probability, however, exact recovery requires further assumptions on diversity of the domains. Later, \cite{HD17} extended this work to non-linear functional relations, yet still assuming additive noise. \cite{PF18} used invariant causal prediction assumption to handle sequentially ordered data where heterogeneity stems from the varying index of the datapoints. \cite{GH17} considered invariance of functional relations between variables and their parents in the linear SCM while the distribution of exogenous noises may vary among the domains. The authors introduced a notion of completeness of causal discovery for this setting and proposed a complete algorithm which is computationally intensive. They also presented a heuristic algorithm that improved upon the complete baseline in computational complexity. \cite{KZ19} (and previously, \cite{KZ17}), proposed a more generalized model for the multi-domain causal discovery problem where the changes in causal mechanisms are modeled by adding deterministic functions from the domain index $C$ to the parameters of causal mechanisms. The variable $C$ itself is also modeled as a random variable in the proposed SCM.  It is assumed that the parameters of the functional relation of variable $V_i$, denoted by $\theta_i(C)$, change independently from the other variables' parameters. However, this assumption may be violated as $\theta_i(C)$ and $\theta_j(C)$ are deterministic functions of the same random variable $C$, and they are unlikely to be independent unless one of them is constant. \cite{ghassami2018multi} proposed a method to identify the causal relation between two random variables in linear SCM where the causal coefficients and/or the distribution of exogenous noises may vary. They also extend this method for a network of variables in linear SCM. The noise model we consider in this work was introduced by \cite{nasr2023counterfactual} as bijective generateion mechanisms (BGM) for the purpose of couterfactual identification, where the graph is assumed to be known; in our analysis, we study a complementary task that is the identification of the graph itself under BGM assumption. \cite{immer2023identifiability} and \cite{strobl2023identifying} showed that location-scale noise model is identifiable, though their analysis is limited to bivariate case with both $X,Y$ being real-valued; On the other hand, our noise model is less restrictive, since the location-scale is a special case in our model, although we require observations from more than one domain to make inferences. An extension of LiNGAM known as multi-group LiNGAM (MGL) is implemented for multi-domain setting. \citep{guo2023causal} introduce the "Causal de Finetti" framework, demonstrating how exchangeable data and the independent causal mechanism assumption can be leveraged to uniquely identify invariant causal structures using multi-domain data. \citep{reizinger2025identifiable} introduce an Identifiable Exchangeable Mechanisms framework that unifies causal discovery, independent component analysis, and causal representation learning by relaxing the conditions for identifying causal structures in exchangeable data. Another method for multi-domain causal discovery is ORION by \citep{mian2023information}, which uses minimum description length (MDL) principles to jointly discover causal networks and sources of heterogeneity across multiple environments that are modeled as hard and soft interventions.

	\subsection*{Contribution} 
	We present a causal discovery algorithm for the multi-domain setting under Markovianity assumption under Bijective generation mechanism (BGM) that generalizes some of the existing noise models including additive noise model \citep{HO09}, LiNGAM \citep{SH06}, post-nonlinear noise model \citep{ZH09}, and location-scale noise model \citep{immer2023identifiability,strobl2023identifying}. Since BGM does not require the exogenous noise to be additive, it captures cases with complex dependencies between the exogenous and endogenous variables, including the case of discrete observables. These dependencies often exist in real-world applications and can cause the methods which are based on independence of the regression residual and the cause to fail. We construct statistical tests to recover the true parent set of a target variable among a set of covariates and prove its soundness under BGM assumptions. Finally, we extend our results to multivariate and mixed-type data in the multi-domain setting.	
	
	The structure of this paper is as follows: In Section \ref{sec:method}, we present our assumptions and the theoretical guarantees for discovery of the parent set, and propose a statistical testing scheme. We also present the extensions of our results to multivariate and mixed-type data setting. In Section \ref{sec:algorithm}, we describe our causal discovery algorithm and heuristics supported by our findings in Section \ref{sec:method}. In Section \ref{sec:experiments} we provide experiment results and compare our method with previous work.
	
 \section{Preliminaries}
Here we provide the background and notation necessary for the other sections.

\begin{defn} A Structural Causal Model is defined by the set of observable variables $\mathbf{V} = \{V_1,V_2,\dots,V_n\}$, a set of mutually independent exogenous variables $\mathbf{E} = \{E_1,E_2,\dots,E_n\}$, a set of functions (i.e., causal mechanisms) $\mathscr{F} = \{f_1,f_2,\dots,f_n\}$, and a probability measure $\mathbb{P}$. In a Markovian SCM, $V_i \gets f_i(\mathbf{PA}_i,E_i)$, where $\mathbf{PA}_i \subseteq \mathbf{V} \setminus \{V_i\}$ is called the parents of $V_i$. Moreover, the causal graph $G$ is constructed over the vertex set $\mathbf{V}$ by adding directed edges from every $V' \in \mathbf{PA}_i$ to $V_i$. We assume that the SCM is non-recursive, i.e., $\mathscr{G}$ is acyclic.
\end{defn}

In this paper, we first restrict ourselves to a bivariate case with the causal diagram $X \to Y$ which indicates that $Y \gets f(X,E)$, and then we extend the results to the multivariate case. Due to Markovianity (a.k.a. causal sufficiency), we assume that $X,E$ are independent random variables. The sets $\mathcal{X}$,$\mathcal{Y},$ and $\mathcal{E}$ denote the support of $X,Y,E$, respectively.
	
There exists a variety of models for multi-domain causal discovery in the literature \citep{jaber2020causal,perry2022causal,huang2019causal,peters2016causal,mooij2020joint,yang2018characterizing,squires2020permutation}. In order to identify the causal structure uniquely, most of these methods not only assume Markovianity, but also require that either the functional relations $f_i$ or the distribution of exogenous variables $P(E)$ are invariant across the different domains. Furthermore, as discussed in the related work, each approach asserts additional assumptions about the noise models, some more restrictive than others.
	
 To model a multi-domain setting, we assume that the distribution of variables in each domain is governed by a probability measure; the set of these probability measures is denoted by  $\mathcal{M} = \{ \mathbb{P}^1, \mathbb{P}^2, \dots, \mathbb{P}^m \}$, where $\mathbb{P}^i$ corresponds to the $i$-th domain. We assume that these probability measures are defined over the same $\sigma$-algebra $\Sigma$ so that for each $\mathbb{P}^i \in \mathcal{M}$, the triple $(\Omega, \Sigma, \mathbb{P}^i)$ is a probability space. Random variables $X,Y,E$ can have different distributions under each probability measure, but the distribution of $Y$ depends on the distributions of $X$ and $E$ through the function $f: \mathcal{X}\times \mathcal{E} \to \mathcal{Y}$ that we assume to be invariant across the domains. For each random variable $V: \Omega \to \mathcal{V}$ (such as $X$ and $Y$) and a value $v\in \mathcal{V}$ (such as $x \in \mathcal{X}$), we denote the distribution under $\mathbb{P}^i \in \mathcal{M}$ as $p^i_V(v)$. Similarly, for a pair of random variables $V,W \in \mathbf{V}$, we denote the distribution of $V$ conditional on $W$ as $p^i_{V|W}(v|w)$.
	
For multivariate functions like $g:\mathcal{A} \times \mathcal{B} \to \mathcal{C}$, and the points $a\in \mathcal{A}, b\in \mathcal{B}$, we write $g(\cdot,b)$ or $g(a,\cdot)$ to denote the functional $g_b: \mathcal{A} \to \mathcal{C}$ and $g_a: \mathcal{B} \to \mathcal{C}$, respectively, which are defined as $g_b(a) = g_a(b) = g(a,b)$. We view conditional probability densities as multivariate functions defined over the support of the random variables.
	
 \section{Identification Result}\label{sec:method}
	
	We assume that the joint density function $p^i_{X,Y}$ exists under each $\mathbb{P}^i \in \mathcal{M}$. In addition, we assume the set of probability measures $\mathcal{M}$ are mutually absolutely continuous, i.e., each subset of the sample space either has zero probability under every $\mathbb{P}^i \in \mathcal{M}$, or it has a positive probability under every $\mathbb{P}^i \in \mathcal{M}$; This assumption is known as positivity, it simplifies the results and the notation, and is standard in the literature \citep{rosenbaum1983central, pearl2009causal, shpitser2008complete, peters2016causal, dawid2010beware, imbens2015causal}. Thus, under positivity assumption, the densities $\{ p^i_{X,Y} \}_{i=1}^m$ would all have the same support. We also assume that the function $f:\mathcal{X}\times \mathcal{E} \to \mathcal{Y}$ to be invariant across the domains.
	
	We assume Markovianity, i.e., causal sufficiency in all domains, that is $E \independent X$ in under every $\mathbb{P}^i \in \mathcal{M}$, or equivalently, $p^i_{E|X}(e|x) = p^i_E(e)$ for all values of the noise $E=e$. This assumption is also known as the principle of independent mechanisms  is commonly imposed in the literature of multi-domain or single-domain causal discovery \citep{scholkopf2012semi, Elements, lemeire2013revisiting}.

	\begin{defn}
		A fixed-cause functional at $x \in \mathcal{X}$ is denoted by $f(x,\cdot)$ which represents the functional relation between the exogenous noise $E$ and the effect variable $Y$, for a specific value of cause, $x$.
	\end{defn}
	
	\begin{table}
		\caption{Noise Models}
		\label{tbl:noisemodels}
		\centering
		\begin{tabular}{lll}
			\toprule
			{\specialcellbold{Model}}                               & {\specialcellbold{Functional Form}} & {\specialcellbold{Fixed-Cause Functional $f(x,\cdot)$}}\\
			\midrule
			Additive noise model & $f(X,E) = g(X) + E$ & Affine: $g(x) + E$ \\
			LiNGAM  & $f(X,E) = \beta.X + E$ & Affine: $\beta.x + E$\\
			Post-nonlinear model  & $f(X,E) = h(g(X) + E)$ & {\specialcell{monotone $\circ$ affine: $h(g(x) + E)$ }}\\
			location-scale noise model  & $f(X,E) = g(X) + h(X)E$ & Affine : $g(x) + h(x)E$\\
			\bottomrule
		\end{tabular}
	\end{table}
	
	\begin{defn}
		A bijective function $g:\mathcal{A} \to \mathcal{B}$ is a diffeomorphism if it is differentiable and also has a differentiable inverse.
	\end{defn}

    Below, we introduce the key assumption we use in this paper.

	\begin{assm}[bijective generation mechanisms] \label{assm:mechanism}
		For every $x \in \mathcal{X}$, we assume that the fixed-cause functional $f(x,\cdot)$ is a diffeomorphism.
	\end{assm}
	
	Notably BGM extends all classical noise models in the Table \ref{tbl:noisemodels}, since they all satisfy this Assumption \ref{assm:mechanism}.
        There are some recent works on learning causal directions by considering additional assumptions on the data generation mechansims. For example, in the CdF theorem (bivariate case) \citep{guo2023causal}, it is assumed that there are two latent random variables corresponding to the causal mechanisms generating the cause and effect (given the cause). However, our approach does not take a Bayesian perspective and does not require such latent variables. Another line of research relies on nonlinear ICA, which often imposes additional assumptions on exogenous noise, such as the ``variability" assumption \citep{hyvarinen2019nonlinear}. In contrast, our work does not require these assumptions. For instance, in \citep{reizinger2023jacobian}, Assumption 2(v) states that the conditional distribution of certain pairs of exogenous noises must follow a von Mises-Fisher distribution.

    We are given $n_i$ i.i.d. samples $\{ (x^i_l,y^i_l) \}_{l=1}^{n_i}$ for each $\mathbb{P}^i \in \mathcal{M}$, and our task is to decide whether $X\to Y$ is a valid causal diagram under BGM.
    
	\subsection{Necessary Conditions Implied by BGM}
	\begin{defn}
		Define $\tilde{Y}_x$ as a $\mathcal{Y}$-valued random variable with the same distribution as $Y$ conditioned on $X=x$, under each $\mathbb{P}^i \in \mathcal{M}$. More precisely, for each measurable subset $u \subset \mathcal{Y}$ and for each $1\leq e\leq m$, we have
		\begin{equation}
		\mathbb{P}^i(\tilde{Y}_x \in u) = \mathbb{P}^i(Y \in u|X = x).
		\end{equation}
		Since we assumed the joint probability density function $p_{X,Y}$ exists, this is equivalent to
		\begin{equation}
		p^i_{\tilde{Y}_x}(y) = p^i_{Y|X}(y|x),
		\end{equation}
		for each $1\leq e\leq m$ and each $y \in \mathcal{Y}$.
	\end{defn}
	
	\begin{defn}[Identical r.v.s] \label{defn:id}
		Consider two $\mathcal{V}$-valued random variables $V$ and $V'$. $V$ and $V'$ are ``\textit{identical}" (denoted by $V \overset{I}{=} V'$) if for every $\mathbb{P}^i \in \mathcal{M}$ and for each $u \subset \mathcal{V}$,
		\begin{equation}
		\mathbb{P}^i(V \in u) = \mathbb{P}^i(V' \in u).
		\end{equation} 
	\end{defn}

	\begin{defn}[Similar r.v.s] \label{defn:sim}
		Let $A$ and $B$ be two $\mathcal{A}$-valued and $\mathcal{B}$-valued random variables, respectively. $A$ is ``\textit{similar}" to $B$ (denoted by $A \sim B$) if and only if there exists a diffeomorphism $g:\mathcal{A} \to \mathcal{B}$ such that $B \overset{I}{=} g(A)$.
	\end{defn}
	
	For any random variable $V$ and diffeomorphism $g$, density function of $W = g(V)$ under $\mathbb{P}^i \in \mathcal{M}$ is $p^i_W(w) = \frac{p^i_V(g^{-1}(w))}{|det(J_g(g^{-1}(w)))|}$, where $J_g$ denotes the Jacobian matrix of the $g$ at each point in the support of $V$. As $g$ is a diffeomorphism, the matrix $J_g$ exists and is non-zero at every point of support of $\mathcal{V}$ \citep{Stark}. Thus, for any pair of similar random variable random variables $A \sim B$ (Definition \ref{defn:sim}), we have
	\begin{equation} \label{eq:A-B}
	\frac{p^1_A(g^{-1}(b))}{p^1_B(b)} =\frac{p^2_A(g^{-1}(b))}{p^2_B(b)} = \dots = \frac{p^m_A(g^{-1}(b))}{p^m_B(b)}.
	\end{equation}
	
	It can be easily shown that in a single-domain setting, any two continuous probability density functions can be transformed into each other by diffeomorphisms \cite{Nelsen}. Therefore, similarity of two random variables is trivially true as long as $|\mathcal{M}|=1$. However, in a multi-domain setting, i.e., $|\mathcal{M}| >1$, similarity of random variables would be a non-trivial relation. Below, we prove a property that holds under BGM which would in turn enable causal discovery.
	
	\begin{prop} \label{prop:simconds}
		Assumption \ref{assm:mechanism} holds  if and only if 
		\begin{equation}
		\forall a,b \in \mathcal{X}: \tilde{Y}_a \sim \tilde{Y}_{b}.
		\end{equation}
		We call this property ``pairwise similarity of conditionals".
	\end{prop}

    To further elaborate on the above, we note that similarity of random variables requires existence of a diffeomorphism that transforms one to the other. In a multi-domain setting, not every two continuous random variables are similar, and therefore, we can test this property. According to Proposition \ref{prop:simconds}, in a bijective generation mechanisms (i.e., Assumption \ref{assm:mechanism}), the variables that represent the conditional distributions of the effect variable $Y$ are all similar.
	
	Proposition \ref{prop:simconds} can be used as a basis to reject $X\to Y$ as follows: if we find $a,b \in \mathcal{X}$ such that $\tilde{Y}_a \not\sim \tilde{Y}_b$, that is $\tilde{Y}_b \not \overset{I}{=} g(\tilde{Y}_a)$ for any diffeomorphism $g: \mathcal{Y} \to \mathcal{Y}$, we can reject $X \to Y$. Note that verifying pairwise similarity of conditionals, even for a single pair $a,b \in \mathcal{X}$, is intractable in this way since there are countless candidates diffeomorphisms that one must consider. In fact, Proposition \ref{prop:simconds} is just an identifiability result and it does not provide a method on how to test for pairwise similarity property. Next, we introduce a practical method for statistically testing pairwise similarity of conditionals.
	
 \subsection{Statistical Tests for Similarity}
	First, we introduce a probabilistic indicator of similarity. Next, we propose an algorithm by which we can measure this indicator using finite samples.
	
	According to Eq. \eqref{eq:A-B}, for any two similar random variables $A$ and $B$ such that $B=g(A)$, the vectors $(p^1_A(g^{-1}(b)),\cdots,p^m_A(g^{-1}(b)))^T$ and $(p^1_B(b),\cdots,p^m_B(b))^T$ are aligned. Based on this observation, we define the following vector representation associated with random variables.

	\begin{defn} \label{defn:Phi}Consider $\mathcal{V}$-valued random variable $V$. The density-vectorization associated with $V$, $\Phi_{V} : \mathcal{V} \to S_{m}$, is defined as
		\begin{equation}
		\Phi_{V}(v) = \frac{h}{\|h\|_1},
		\end{equation}
		where $S_m$ is the simplex in $m$-dimensional space and
		\begin{equation}
		h = (p^1_{V}(v), p^2_{V}(v), \dots, p^m_{V}(v))^T,
		\end{equation}
		for every $v \in \mathcal{V}$.
	\end{defn}

    In words, the density-vectorization operator takes as input a point in the support of $V$, concatenates the density of $V$ across the domains into a vector of $m$ entries, and then normalizes it so that they sum up to one. The reason for using this operator is to summarize the relative density of $V$ in each point of the support, since this quantity turns out to be an important invariant of this setting that we leverage for identification.
    
	\begin{figure}
		\centering
		\includegraphics[width=0.7\linewidth]{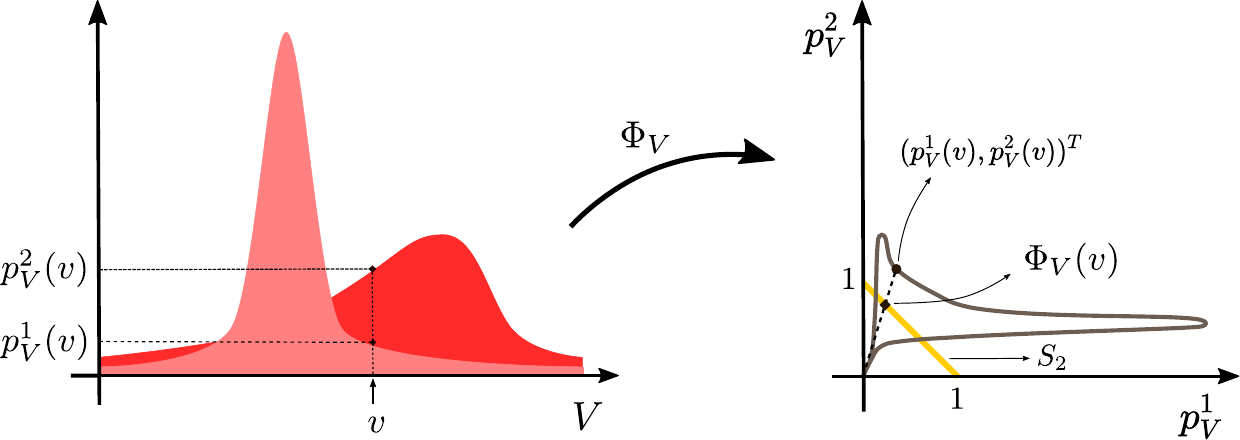}
		\caption{Left: Density functions $p^1_V,p^2_V$ for random variable $V$. Values of density functions at point $v$ are also shown in this plot. Right: Points on the gray curve are vectorization of values of density functions $p^1_V,p^2_V$ for each point $v$ in $\mathbb{R}$. Projecting the density vector $(p^1_V(v),p^2_V(v))^T$ on the simplex $S_2$ yields $\Phi_V(v)$ which is shown on the plot.}
		\label{fig:phi}
	\end{figure}

	\begin{defn}
		Random variable $\Psi_{V}:\Omega \to S_m$ is called the ``special random variable" associated with $V$ and it is defined as $\Psi_{V} := \Phi_{V}(V)$. Note that $\Phi_V$ is a deterministic function, but $\Psi_V$ is a random variable as it is a deterministic image of $V$.
	\end{defn}
	
	\begin{prop} \label{prop:sim2id}
		Consider $A$ and $B$, two $\mathcal{A}$-valued and $\mathcal{B}$-valued random variables, respectively. If $A \sim B$, then $\Psi_{A}  \overset{I}{=} \Psi_{B}$.
	\end{prop}
	
	In words, if two random variables are similar, then the special random variables associated with them are identical. Note that this result does not introduce an equivalent condition for similarity of random variables; instead, it is a necessary condition, and can be used to reject the hypothesis of similarity whenever the latter condition about special random variables is violated\footnote{We conjecture that $\Psi_{A}  \overset{I}{=} \Psi_{B}$ is a sufficient condition for $A$ and $B$ to be similar. We could not generate any counter-examples, but could not prove the statement either.}. That being said, in case the distribution of variables does not change between the domains, then the density-vectorization operator collapses all points of the support to the point $\underbrace{\langle \frac{1}{m},\frac{1}{m}, \dots, \frac{1}{m}\rangle }_{m \text{ times}}$. This implies that $\Psi_{V}$ is a constant, thus would be independent of all random variables. Therefore, if the distribution of variables in different domains are identical then the tests would be uninformative since no hypothesis may be rejected.

	Proposition \ref{prop:simconds} implies that Assumption \ref{assm:mechanism} is equivalent to pairwise similarity of conditionals. Moreover, according to Proposition \ref{prop:sim2id}, Assumption \ref{assm:mechanism} implies that for every pair of values $a,b \in \mathcal{X}$, the special random variables associated with $\tilde{Y}_a$ and $\tilde{Y}_b$ are identical. Hence, if there exists $a,b \in \mathcal{X}$ for which $\Psi_{\tilde{Y}_a} \overset{I}{\not =} \Psi_{\tilde{Y}_b}$, then $X \not\to Y$ under our assumptions. Motivated by this observation, we design a statistical test to reject pairwise similarity, thus, the causal diagram $X\to Y$ under BGM.
	
	\begin{defn} \label{defn:gamma}
		Let $\Gamma_{X \to Y}$ be an $S_m$-valued random variable obtained by taking a random sample $x \in \mathcal{X}$ according to the distribution of $X$ (in each domain), and then drawing a random sample from $\Psi_{\tilde{Y}_x}$. More formally,
		\begin{equation}
		\Gamma_{X \to Y} := \Psi_{\tilde{Y}_{X}} = \Phi_{\tilde{Y}_{X}}(\tilde{Y}_{X}).
		\end{equation}
	\end{defn}
	
	\begin{prop} \label{thm:theone}
		For every $a,b \in \mathcal{X}$, $\Psi_{\tilde{Y}_a}  \overset{I}{=} \Psi_{\tilde{Y}_b}$ if and only if
		\begin{equation}
		\Gamma_{X \to Y} \independent X, \text{ under each }  \mathbb{P}^i \in \mathcal{M}.
		\end{equation}
	\end{prop}
	
	\begin{cor}
		Combining the latter with Propositions \ref{prop:simconds} and \ref{prop:sim2id} implies that we can reject the causal direction $X \to Y$ if data does not statistically admit the above independence relation under some $\mathbb{P}^i \in \mathcal{M}$. Since if $\Gamma_{X \to Y} \not\independent X, \text{ under some }  \mathbb{P}^i \in \mathcal{M}$, then there would be some $a,b\in \mathcal{X}$ such that $\Psi_{\tilde{Y}_a}  \overset{I}{=} \Psi_{\tilde{Y}_b}$ does not hold. Thus, based on Proposition \ref{prop:sim2id}, $\tilde{Y}_a$ and $\tilde{Y}_b$ are not similar which violates Assumption \ref{assm:mechanism} according to Proposition \ref{prop:simconds}.
	\end{cor}
	
	Based on the above corollary, we state our main result as follows:
	
	\begin{thm} \label{theorem}
		Under Assumption \ref{assm:mechanism} and for continuous random variables $X$ and $Y$, the causal diagram $X\to Y$ can be rejected if $\Gamma_{X \to Y} \not\independent X, \text{ for some }  \mathbb{P}^i \in \mathcal{M}$.
	\end{thm}
	
	In what follows, we describe two extensions to our model. First, we discuss how our results extends to discrete case. Second, we discuss the extension to network of multiple variables. In these extensions, we modify both some of our assumptions in Section \ref{sec:method} as well as our theoretical results to cover the extended settings\footnote{Our finding could be expressed in measure theoretic language where the continuous and discrete cases (as well as the cases which are neither continuous nor discrete) can be unified. To make the presentation accessible to a larger audience, we decided to adopt the current presentation which avoids complex notation and technical discussions in measure theoretic formalism.}.
	\subsection{Discrete Case} \label{sec:discrete}
	Assume that $E$, and consequently $Y$, are discrete random variables\footnote{Since we assume that $f(x,\cdot)$ is a bijective function from $\mathcal{E}$ to $\mathcal{Y}$, $Y$ should take discrete values.}. Continuity and differentiability are not defined for mappings between discrete sets. Thus, we replace Assumption \ref{assm:mechanism} with the following assumption:
	\begin{assm}\label{assm:mechanismdiscrete}
		For every $x \in \mathcal{X}$, the fixed-cause functional $f(x,\cdot)$ is bijective.
	\end{assm}
	
	Akin to Definition \ref{defn:sim}, discrete variables $A,B$ are similar (denote it by $A \sim B$), if and only if there exists a bijection $g$ such that $B \overset{I}{=} g(A)$. 
	
	\begin{prop} \label{theorem_discrete}
    	For discrete-valued $E$, under Assumption \ref{assm:mechanismdiscrete}, there is no causal direction from $X$ to $Y$ if $\Gamma_{X \to Y} \not\independent X, \text{ for some }  \mathbb{P}^i \in \mathcal{M}$.
	\end{prop}

 In words, we can still test the causal diagram $X\to Y$ using the same exact method described for the continuous case.
 
	\subsection{Multivariate Case} \label{sec:mult}
 
	In this subsection, we extend our results to a setting with more than two observed variables by extending the Assumption \ref{assm:mechanism} to the multivariate case as follows.
	
	\begin{assm} \label{assm:mechanismmult}
		For every value of $V_i$'s parents like $pa_i$, we assume that the fixed-cause functional $f_i(pa_i,\cdot)$ is a diffeomorphism.
	\end{assm}
	
	Notably Assumption \ref{assm:mechanismmult} is an extension of Assumption \ref{assm:mechanism} for each endogenous variable.
	
	\begin{prop} \label{theorem_mult}
	    Under Assumption \ref{assm:mechanismmult},
		if for each variable $V_i$, its conditional density function given any subset of variables $\mathbf{S} \subset \mathbf{V}$ is continuous in all domains, then $\mathbf{S} \neq \mathbf{PA}_i$ if
		\begin{equation} \label{eq:test}
		    \Gamma_{\mathbf{PA}_i \to V_i} \not\independent \mathbf{PA}_i, \text{ for some }  \mathbb{P}_i \in \mathcal{M}.
		\end{equation}
	\end{prop}
	
	In words, the conditional independence test in Eq. \ref{eq:test} can be used as a basis for rejecting the hypothesis of a certain set being the parent set of $V_i$. Similarly, note that this result can be extended to mixed-type (continuous or discrete) multivariable case. In what follows, we construct the discovery routine based on our findings.
	
	\section{Algorithm} \label{sec:algorithm}
	We assume that in each domain, there are $n_e$ i.i.d. samples drawn from $\mathbb{P}_i \in \mathcal{M}$. Denote these observations as $\{ (x^i_l,y^i_l) \}_{l=1}^{n_e}$.\footnote{To cover the discrete extension from Subsection \ref{sec:discrete}, we use the general term ``distribution" to refer density functions in continuous settings and probability mass functions in discrete settings.}
	\subsection{Sampling From $\Gamma_{X \to Y}$} \label{sec:gamma}
	In Subsection \ref{sec:mult} we propose the algorithm in the general multivariate case. If the conditional distribution functions $\{ p^i_{Y|X} \}_{i=1}^m$ were available, we could obtain random samples from $\Gamma_{X \to Y}$ under each of the probability measure as followings. Note that $\tilde{Y}_{X}$ (Definition \ref{defn:gamma}) and $Y$ are identically distributed, since the subscript $X$ is randomized. For a datapoint $(x,y)$ from say domain $e$, we form the vector $w = (p^1_{Y|X}(y|x), p^2_{Y|X}(y|x), \dots, p^i_{Y|X}(y|x))^T$ and project it on $S_m$ (the simplex in $m$ dimensions) to obtain $\gamma := \frac{w}{\|w\|_1}$, which would be identically distributed as $\Gamma_{X \to Y}$ under $\mathbb{P}^i$. When the true conditional distribution functions are not available, assuming we have enough samples from the joint observations of $X$ and $Y$, we estimate the conditional distribution function $p^i_{Y|X}$ in each of the domains: $\hat{p}^i_{Y|X}$.\footnote{We performed this estimation using \textsc{np} package in R \cite{np} in our implementation, and the code is available in the supplementary material.} Using these estimated conditionals, we can obtain joint samples of $(\Gamma_{X \to Y}, Y)$ as described above for each domain.
	
	Upon consistent estimation of the conditional distributions, this sampling routine yields ``approximately accurately distributed'' samples as $(\gamma,x)$. As the size of data increases (in all domains), we obtain more accurate samples of $(\Gamma_{X \to Y},X)$, as we can estimate the conditional distributions more accurately across the domains.

	\subsection{Inference} \label{inference}
	As described in Subsection \ref{sec:gamma}, we can obtain samples of $\Gamma_{X \to Y}$ in each domain. Using these samples, we perform an independence test to evaluate $\Gamma_{X \to Y} \independent X \text{ under each } \mathbb{P}^i$.\footnote{In our implementation, we used d-variable HSIC test \cite{HSIC} provided in \text{dHSIC} package in R.} According to Theorem \ref{theorem}, we shall reject $X\rightarrow Y$ if this independence relation is rejected under any probability measure $\mathbb{P}^i \in \mathcal{M}$. To aggregate the results of these $m$ independence tests (one in each of the $m$ domains), we consider the minimum $p$-values of the independence tests performed in each of the domains. This ensure that the output of the aggregation is small whenever the independence relation is rejected in at least one domain. 
	
	Without further assumptions about the data generation mechanism, the true causal structure is identifiable up to Markov equivalence classes (skeleton\footnote{The skeleton of a directed graph $G$ is an undirected graph which does not take the direction of edges of $G$ into account.} and v-structures of the causal graph) \citep{Elements}. As we are mainly concerned with orienting the undirected edges of the skeleton graph, we seek to find the set of parents of the nodes in the graph. Let $V$ denote a variable in the SCM, and $\mathbf{A}$, the set of all the variables adjacent to $V$. Clearly, parents of $V$ denoted by $PA(V)$ (or for short $\mathbf{PA}$) is a subset of $\mathbf{A}$. Let $L(\mathbf{S})$ be the minimum p-value of testing $\Gamma_{\mathbf{S} \to V} \independent V$ over all of the domains. If $L(\mathbf{S})$ is lower than some threshold $c$, then there is enough evidence that at least in one domain, the independence relation $\Gamma_{\mathbf{S} \to V} \independent V$ is violated. As a result of Theorem \ref{theorem}, if $L(\mathbf{S})<c$, we conclude that $\mathbf{S} \neq \mathbf{PA}$. In order to obtain a single subset of $\mathbf{A}$ as the inferred parent set, we propose the following methods:
	
	\noindent \textbf{H1}: Compute $L(\mathbf{S})$ for all $\mathbf{S} \subset \mathbf{A}$. To ensure that $\mathbf{PA}$ is contained in the output, return 
	\begin{equation} \label{eq:H1}
	\mathbf{\hat{PA}} := \bigcup_{\mathbf{S} \subset \mathbf{A}: L(\mathbf{S}) > c  } \mathbf{S}.
	\end{equation}
	
	\noindent \textbf{H2}: If we know the size of the parent set apriori or have a bound on it, it suffices to explore subsets with that size or up to that bound, and return the subset with maximum value of $L$. 

 \section{Experiments} \label{sec:experiments}

    Next, we evaluate the performance of proposed methods, and we explore use of our algorithms in synthetic and real data, and compare it with related work.
	\subsection{Synthetic Data}
\begin{figure}
		\centering
		\includegraphics[width=0.6\linewidth]{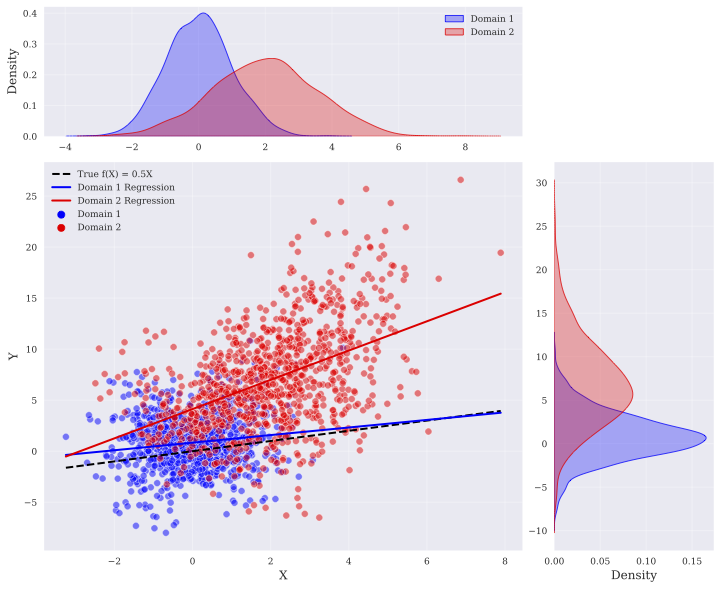}
		\caption{1000 datapoints in two domains, the marginals of $X$ and $Y$, the regression lines in each domain, and the true invariant component $f$.}
		\label{fig:bivariate-data}
	\end{figure}

	We used a heteroscedastic noise model to generate our synthetic data in two domains. In this model, variable $Y$ is determined by the value of its parent $X$ and an exogenous noise $E$ by the following equation
	\begin{equation} \label{hetmodel}
	    Y = f(X) + g(X)E,
	\end{equation}
	where we assumed $f(X) := \alpha^TX$, and $g(X)$ was randomly selected with equal probability from $\sqrt{|\beta^TX| + 1}$ or $\log(|\beta^TX| + 2)$, respectively. We focused on the setting discussed in Subsection \ref{inference}, in which we observe variables $\mathbf{A} \cup \{V\}$, and the value of each variable is determined by the value of its parents according to Eq. \eqref{hetmodel}. The coefficients  $\alpha$ and $\beta$ for each structural equation were drawn randomly from $\mathcal{N}(0,0.05)$ and $\textit{unif}([-2,-1] \cup [1,2])$, respectively. We considered two domains and in both domains, the exogenous noise was assumed normal. Let $\mu_e$ and $\sigma_e$ be the mean and the standard deviation of exogenous noise in domain $e \in \{1,2\}$, respectively. We drew $\mu_1$ randomly from $\mathcal{N}(0,1)$ and $\sigma_1$ was set to 2. We also set the parameters in the second domain to $\mu_2 = \mu_1 + \text{\textit{unif}}(\{-3,3\})$ and $\sigma_2 = \sigma_1\text{\textit{unif}}([\frac{2}{3},\frac{3}{2}])$. An instance of the data in bivariate case is shown in Figure \ref{fig:bivariate-data}.

	We compared the proposed methods with ICP \citep{JP16}, NLICP \citep{HD17}, MC \citep{ghassami2018multi},  IB \citep{ghassami2018multi}, LRE \citep{GH17}, CD-NOD \citep{KZ19}, CdF \citep{guo2023causal}, and multi-group LiNGAM (MGL) \citep{shimizu2012joint} which have been proposed previously for the multi-domain setting. ICP method assumes that the target variable is a linear function of a subset of predictors, where the coefficients may change across the domains while the distribution of additive exogenous noise is assumed to be invariant. Non-linear ICP allows the functional relation to be non-linear. LRE considers invariant linear functional relations while the distribution of additive exogenous noise might change among the domains. IB and MC extend LRE so that neither distribution of noise nor the linear functional relation is assumed to be invariant. CD-NOD recovers causal relations based on the assumption of independent changes of conditional probability $\mathbb{P}($cause$|$effect$)$. In our experiments, we also considered LiNGAM \citep{SH06} which is a causal discovery algorithm in single domain setting and assumes linear functional relations with additive non-Gaussian noise.
	
	We evaluated the performance of our algorithms as well as previous work in two cases (each case comprised of 100 instances of synthetic dataset). We tuned the hyper-parameters of all algorithms (e.g. $c$ for our algorithm) with a training dataset before each experiment, and evaluated the performance on another testing dataset. In both experiments, H2 was initially fed with the size of true parent set. The code is made available here: \href{https://github.com/jalaldoust/MD-BGM}{https://github.com/jalaldoust/MD-BGM}.

 \begin{table}[t]
		
	    \centering
	    \small
		\begin{tabular} {ccccccccccc}
			\toprule
			{\specialcellbold{Alg.}}                               & {H1} & {ICP } & {NLICP } & {LiNGAM } & {MC  } & {IB  } & {LRE } & {CD-NOD } & {CdF} & {MGL}\\
			\midrule
			{\specialcellbold{Acc.}}                               & {82\%} & {43\%} & {51\%} & {66\%} & {62\%} & {59\%} & {43\%} & {10\%} & {49\%} & {63\%}\\
			\bottomrule
		\end{tabular}
  \caption{Bivariate case: Accuracy of algorithms on detection of causal direction between two variables. }
  \label{tbl:comparison_bi}
	\end{table}

	\noindent \textbf{Bivariate case.} In this case, $|\mathbf{A}|=1$. The adjacent variable was set as the parent of $V$ with probability $\frac{1}{2}$. We generated $1000$ samples in each of the domains. Accuracy of the algorithms are reported in Table \ref{tbl:comparison_bi}. The proposed method (H1) indeed achieves the highest accuracy. It is worth to note that since we used only 100 instances of the problem, the standard deviation of the accuracies is no more than $5\%$, which still deems our algorithm as the most accurate with high confidence.

\begin{figure}[t]
  \includegraphics[width=\linewidth]{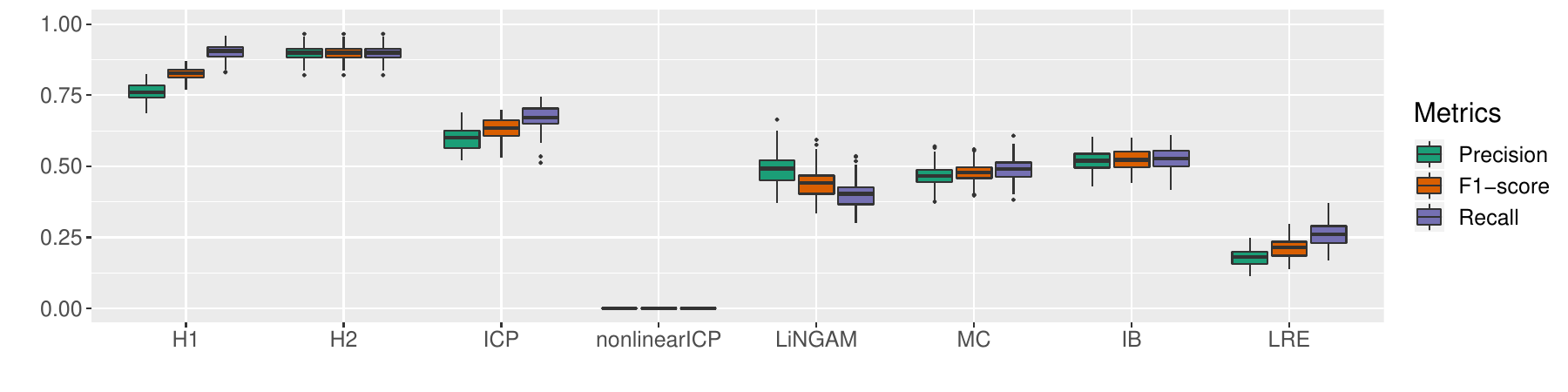}
  \caption{Multivariate case: Performance of our methods 
  (H1 and H2) as well as previous work.}
  \label{fig:comparison_mult}
\end{figure}

	\noindent \textbf{Multivariate case.} In this case, $|\mathbf{A}|$ was selected uniformly at random from $\{ 2,3,4,5 \}$, and we picked $|\mathbf{PA}|$ uniformly from $\{ 1, 2, ..., |\mathbf{A}| \}$. We considered the fully connected skeleton among the variables $\mathbf{A} \cup \{V\}$ and the parents were determined according to a random topological order. We generated $10000$ samples in each of the domains and evaluated Precision, Recall, and F1-score for each of the algorithms. We repeated this procedure 100 times. Figure \ref{fig:comparison_mult} depicts the performances of the various algorithms we compared. H1 outperformed all other algorithms with 20\% margin of F1-score. H2  which has the advantage of prior knowledge about the size of parent set, has 12\% improvement in precision over H1. (Please note that CD-NOD algorithm does not return any output in a reasonable time for the size of parents greater than one. Thus, its performance is reported only in the bivariate case).  
	

	\subsection{Real World Data}
	\textbf{CollegeDistance dataset.} We considered educational attainment data \citep{educ} which was collected from approximately 1100 high-school students. The data contains 13 features including gender, race, base year composite test score, family income, etc. As in \cite{JP16} and \cite{GH17}, we split the observations into two groups (treated as two domains). This is done based on the distance to the closest 4-year college where the first group contains students within the 10 miles radius. The target variable is the number of years of education. We run our method for $100$ trials and in about $80\%$ of the executions, the set with greatest value of $L(\mathbf{S})$ was $\mathbf{S} = \{$race$\}$. Thus, this variable was returned as the parent set of the target variable. This variable along with three other candidates was also selected in \cite{GH17} as the parent set.

	\noindent \textbf{Adult dataset.} This dataset is available at UCI Machine Learning Repository \cite{Adult} and contains census information. We pre-processed the data by filtering  or transforming some of the features before feeding them to our causal discovery algorithm. We chose sex as the domain variable, and working hour per week as the target variable. We considered the following features: age,  race, marital status, level of education, level of income, work class, and country. Among these variables, the maximum value of $L(\mathbf{S})$ was achieved by $\mathbf{S} = \{ \text{country} \}$. This is not surprising since working policy in each country has a direct impact on the working hour per week.





    \subsection{Limitations}
    Our algorithm has few practical limitations. First, accurately estimating the density in each environment may require large amounts of data, and the minimum required data grows exponentially with the dimensionality of the variables unless additional assumptions are imposed.
Second, our framework requires domain indices to be integers, and the causal discovery task depends on having sufficient data in all domains. Consequently, our method is most effective when the number of domains is small relative to the dataset size.
Third, the computational cost is significant due to the combinatorial search over parent sets, computation of test statistics, and execution of independence tests.
A theoretical limitation of our approach is that we do not claim completeness of the discovery algorithm—while we guarantee soundness (i.e., once a non-parent is identified, it is truly a non-parent), we do not ensure that all non-parents are rejected. Since completeness is not the focus of this work, the discussion of necessary faithfulness assumptions remains an open question.
    
    
	\section{Conclusion}
	We studied causal discovery problem in multi-domain setting with invariant causal relations and varying noise distribution. We presented an identifiablity result assuming bijective generation mechanisms, which is less restrictive than previous approaches, and holds in several classical models. We defined \textit{similarity} relation between random variables in settings with multiple probability measures and we introduced an independence relation which necessarily holds in case of causation. We compared the performance of our approach with previous work and in our experiments our method out performed them with a high margin. Future work includes extension of results to non-stationary time-series and applications in sequential decision-making.

\newpage
\bibliography{bib}

\newpage
\appendix

\begin{center}
    \textbf{\Large Appendix}
\end{center}
	
	\begin{rem} \label{rem:diffs}
		
		Suppose $g,f:\mathcal{A} \to \mathcal{B}$ are diffeomorphisms. Then $g^{-1}$ and $g$o$f$ are also diffeomorphisms.
		
	\end{rem}
	
	\begin{lem}\label{lem:equivalence}
		Similarity is an equivalence relation.
		
		\begin{proof}
			
			We should check the three properties of equivalence relations. Consider random variables $A$, $B$, and $C$ taking value in $\mathcal{A}$, $\mathcal{B}$, and $\mathcal{C}$, respectively. Assume that $A\sim B$ and $B\sim C$.\\
			
			\textit{Claim: $\sim$ is reflective.}
			
			Define $g: \mathcal{A} \to \mathcal{A}$ such that $ \forall x \in \mathcal{A}: g(x) = x $. It is bijective and continuously differentiable. Hence, it is a diffeomorphism. By Definition \ref{defn:sim}, $A\sim A$.\\
			
			\textit{Claim: $\sim $ is symmetric.}
			
			As $A \sim B$, from Definition \ref{defn:sim}, there exists a diffeomorphism $g$ such that $g(A) \overset{I}{=} B$. From Definition \ref{defn:id}, for every $e$ and every $u \subset \mathcal{B}$
			\begin{equation}
			    \mathbb{P}^i(g(A) \in u) = \mathbb{P}^i(B \in u).
			\end{equation}
            Note that $g$ is a diffeomorphism, so $g^{-1}$ exists. Let $w $ be an arbitrary subset of $\mathcal{A}$. Let $u = g(w)$, hence $w = g^{-1}(u)$.
            \begin{align}
                \mathbb{P}^i(A \in w)
                &= \mathbb{P}^i(A \in g^{-1}(u))\\
                &= \mathbb{P}^i(g(A) \in u)\\
                &= \mathbb{P}^i(B \in u)\\
                &= \mathbb{P}^i(g^{-1}(B) \in g^{-1}(u))\\
                &= \mathbb{P}^i(g^{-1}(B) \in w).
            \end{align}
			Form Remark \ref{rem:diffs}, we know that $g^{-1}$ is a diffeomorphims. As the choice of $w$ was arbitrary, from Definitions \ref{defn:id} and \ref{defn:sim}, we have 
			\begin{equation}
			    B \sim A.
			\end{equation}
			
			\textit{Claim: $\sim$ is transitive.}
			
			Similarly, for every $e$ and every $u \subset \mathcal{B}$
			\begin{equation}
			    \mathbb{P}^i(g(A) \in u) = \mathbb{P}^i(B \in u).
			\end{equation}
			Also, for every $e$ and every $q \subset \mathcal{C}$,
			 \begin{equation}
			    \mathbb{P}^i(f(B) \in q) = \mathbb{P}^i(C \in q).
			\end{equation}
		    Note that both $g$ and $f$ are diffeomorphisms. Thus, they are invertible. Consider $q \subset \mathcal{C}$ arbitrarily. Let $u = f^{-1}(q)$ and $w = g^{-1}(u)$. 
			\begin{align}
			    \mathbb{P}^i((f \circ g)(A) \in q)
			    &= \mathbb{P}^i(g(A) \in f^{-1}(q))\\
			    &= \mathbb{P}^i(g(A) \in u)\\
                &= \mathbb{P}^i(B \in u)\\
                &= \mathbb{P}^i(f(B) \in f(u))\\
                &= \mathbb{P}^i(f(B) \in q)\\
                &= \mathbb{P}^i(C \in q)
            \end{align}
            From Remark \ref{rem:diffs}, we know that $f \circ g$ is a diffeomorphism. Therefore,
            \begin{equation}
                A \sim C.
            \end{equation}
		\end{proof}
	\end{lem}
	
	\subsection*{ Proof of Proposition
		\ref{prop:simconds}} 	
	\textit{Claim: Assumption \ref{assm:mechanism} holds $\Rightarrow \forall a,b \in \mathcal{X}: \tilde{Y}_a \sim \tilde{Y}_b$. }
	
	Fix the domain variable $e$ and consider $x \in \mathcal{X}$ arbitrarily. As a result of Assumption \ref{assm:mechanism}, $f(x,\cdot)$ is invertible, and we have
	\begin{equation}
	    p^i_{(Y|X)}(y|x) = p^i_{(E|X)}(f(x,\cdot)^{-1}(y)|x) = p^i_{E}(f(x,\cdot)^{-1}(y)),
	\end{equation}
	where the first equation  is because of the model $Y = f(X,E)$, and the second one is due to $X \independent E$.
	
	Let $g := f(x,\cdot)$ which is a diffeomorphism, according to Assumption \ref{assm:mechanism}. As $Y = f(X,E)$, we have
	\begin{equation}
	    \tilde{Y}_x \overset{I}{=} g(E).
	\end{equation}
	Note that $\tilde{Y}_x$ is a random variable which has the same distribution as $Y$ given $X=x$. According to Definition \ref{defn:sim},
	\begin{equation}
	     \forall x\in \mathcal{X}: E\sim \tilde{Y}_x.
	\end{equation}
	By Lemma \ref{lem:equivalence}, similarity is an equivalence relation, so we have 
	\begin{equation}
	    \forall a,b \in \mathcal{X}: \tilde{Y}_a \sim \tilde{Y}_b.
	\end{equation}
	
	\textit{Claim: $\forall a,b \in \mathcal{X}: \tilde{Y}_a \sim \tilde{Y}_b \Rightarrow$ Assumption \ref{assm:mechanism} holds.}
	
	Fix arbitrary $x_0 \in \mathcal{X}$. Define $\tilde{E}$ an independent random variable which is identically distributed as $\tilde{Y}_{x_0}$ ($E \overset{I}{=} \tilde{Y}_{x_0}$). Define $\tilde{f}$ so that $\tilde{f}(x,\cdot) := g_x$ where $g_x$ is the diffeomorphism which forms the similarity between $\tilde{Y}_x$ and $\tilde{Y}_{x_0}$ (Definition \ref{assm:mechanism}). Assumption \ref{assm:mechanism} holds for $\tilde{f}$.

	\subsection*{Proof of Proposition \ref{prop:sim2id}}
	
	Fix the domain index $e$ and consider $s \subset S_m$ arbitrarily. Note that the events $\Psi_A \in s$ and $\Psi_B \in s$ should both be measurable under $\mathbb{P}^i$, as we intend to show 
	\begin{equation}
	    \mathbb{P}^i(\Psi_A \in s) = \mathbb{P}^i(\Psi_B \in s),
	\end{equation}
	which results in $\Psi_A \overset{I}{=} \Psi_B$ (Definition \ref{defn:id}).
	
	Let $\alpha = \Phi^{-1}_A(s)$ and $\beta = \Phi^{-1}_B(s)$. Note that $\alpha \subset \mathcal{A}$ and $\beta \subset \mathcal{B}$. As $A \sim B$, according to Definition \ref{defn:sim}
    \begin{equation}
        \exists \text{ diffeomorphism } g:\mathcal{A} \to \mathcal{B} \text{ such that } g(A) \overset{I}{=} B
    \end{equation}
    Consider $a \in \alpha$ arbitrarily. Let $b = g(a)$. We have
    \begin{align}
        \Phi_B(b) &= c.(p^1_B(b),p^2_B(b),\dots,p^m_B(b)) \label{phidefn}\\
        &= c.\frac{1}{|det(J_g(a))|}.(p^1_A(a),p^2_A(a),\dots,p^m_A(a)) \label{thatremark}\\
	    &= c'.(p^1_A(a),p^2_A(a),\dots,p^m_A(a)),
    \end{align}
	where \eqref{phidefn} holds according to Definition \ref{defn:Phi} and \eqref{thatremark} holds due to \eqref{eq:A-B}. 
	
	As we defined $\alpha$, we have
	\begin{equation}
	\Phi_A(a) = c''.(p^1_A(a),p^2_A(a),\dots,p^m_A(a)) \in s.
	\end{equation}
	Note that we also have
	\begin{equation}
	    \Phi_B(b) = c'.(p^1_A(a),p^2_A(a),\dots,p^m_A(a)) \in s.
	\end{equation}
	Thus, $\Phi_B(b)$ and $\Phi_A(a)$ are proportional with factor $\frac{c'}{c''}$. As they are both in $S_m$, this is possible only if $c'' = c'$ and $\Phi_B(b) = \Phi_A(a)$. Because the choice of $a$ was arbitrary, we have 
	\begin{equation}
	    g(\alpha) \subset \beta.
	\end{equation}
	According to Lemma \ref{lem:equivalence}, $A\sim B \Rightarrow B\sim A$. Therefore, with the same arguments in the reverse direction, we get \begin{equation}
	    g^{-1}(\beta) \subset \alpha,
	\end{equation}
	which results in
	\begin{equation}
	    g(\alpha) = \beta.
	\end{equation}
	Finally,
	\begin{align}
	    \mathbb{P}^i(\Psi_A \in s)
	    &= \mathbb{P}^i(A \in \alpha)\\
	    &= \mathbb{P}^i(g(A) \in g(\alpha))\\
	    &= \mathbb{P}^i(g(A) \in \beta)\\
	    &= \mathbb{P}^i(B \in \beta) \label{30}\\
	    &= \mathbb{P}^i(\Psi_B \in s),
	\end{align}
	where \eqref{30} follows from $g(A) \overset{I}{=} B$ which is imposed by $A \sim B$.
	
	\subsection*{Proof of Proposition \ref{thm:theone}}
	
	\textit{Claim: If for every $a,b \in \mathcal{X}$, $\Psi_{\tilde{Y}_a}  \overset{I}{=} \Psi_{\tilde{Y}_b}$, then
		$\Gamma_{X \to Y} \independent X$, under each   $\mathbb{P}^i \in \mathcal{M}$.}
	
	Fix the domain index $e$. Consider any $s \subset S_m$ arbitrarily. Consider $a,b \in \mathcal{X}$ arbitrarily. Then,
    \begin{align}
        \mathbb{P}^i(\Gamma_{X \to Y} \in s| X=a)
        &= \mathbb{P}^i(\Psi_{\tilde{Y}_X} \in s| X=a) \label{32} \\ 
        &= \mathbb{P}^i(\Psi_{\tilde{Y}_a} \in s) \label{33} \\
        &= \mathbb{P}^i(\Psi_{\tilde{Y}_b} \in s) \label{34} \\
        &= \mathbb{P}^i(\Psi_{\tilde{Y}_X} \in s| X=b)\\
        &= \mathbb{P}^i(\Gamma_{X \to Y} \in s| X=b), 
    \end{align}
    where \eqref{32} is from Definition \ref{defn:gamma} and \eqref{33} is from imposing the condition $X = a$. Moreover, \eqref{34} is from the assumption $\Psi_{\tilde{Y}_a}  \overset{I}{=} \Psi_{\tilde{Y}_b}$ and Definition \ref{defn:id}.
    As the distribution of $\Gamma_{X \to Y}$ is invariant for every condition on $X$, we conclude $\Gamma_{X \to Y} \independent X$ under $\mathbb{P}^i$ and as the choice of $e$ was arbitrary, this holds for every $e$.
	
	\textit{Claim: If $\Gamma_{X \to Y} \independent X$, under each   $\mathbb{P}^i \in \mathcal{M}$, then for every $a,b \in \mathcal{X}$, $\Psi_{\tilde{Y}_a}  \overset{I}{=} \Psi_{\tilde{Y}_b}$.}
	
	Fix the domain index $e$. Consider any $s \subset S_m$ arbitrarily. Consider $a,b \in \mathcal{X}$ arbitrarily. Then,
	\begin{align}
	    \mathbb{P}^i(\Psi_{\tilde{Y}_a} \in s)
	    &= \mathbb{P}^i(\Psi_{\tilde{Y}_X} \in s| X=a)\\
        &= \mathbb{P}^i(\Gamma_{X \to Y} \in s| X=a)\\
        &=\mathbb{P}^i(\Gamma_{X \to Y} \in s| X=b)\\
        &=\mathbb{P}^i(\Psi_{\tilde{Y}_b} \in s).
    \end{align}
    Due to arbitrary choice of $s$, according to Definition \ref{defn:id}, we have 
    \begin{equation}
        \Psi_{\tilde{Y}_a} \overset{I}{=} \Psi_{\tilde{Y}_b}
    \end{equation}
    
    \subsection*{Proof of Theorem \ref{theorem}}
    
    In continuous case, we have Assumption \ref{assm:mechanism}, along with other technical assumptions mentioned in the Section \ref{sec:method}. Proposition \ref{prop:simconds} states that under this set of assumptions, if $X$ causes $Y$ in our model, we have
    \begin{equation}
        \forall a,b \in \mathcal{X}: \tilde{Y}_a \sim \tilde{Y}_b.
    \end{equation}
	Proposition \ref{prop:sim2id} states that for every two similar random variables $A \sim B$, the corresponding special random variables are identical, i.e., 
	\begin{equation}
	    \Psi_A \overset{I}{=} \Psi_B.
	\end{equation}
	Being identical means that they have the same distribution function under every measure $\mathbb{P}^i \in \mathcal{M}$. From these two results, we can imply the following condition from our set of assumptions, in case of $X$ causing $Y$ in our model.
	\begin{equation} \label{condition}
	    \forall a,b \in \mathcal{X}: \Psi_{\tilde{Y}_a} \overset{I}{=} \Psi_{\tilde{Y}_b}.
	\end{equation}
	Proposition \ref{thm:theone} states that from the condition \eqref{condition}, we can imply the following independence criterion in all domains (i.e., for every $\mathbb{P}^i \in \mathcal{M}$):
	\begin{equation}
	    \Gamma_{X\to Y} \independent X.
	\end{equation}
	Therefore, our set of assumptions implies that if the causal direction is from $X$ to $Y$, then the above independence should hold in all domains. As a result, violation of this independence criterion in at least one of the domains, rejects the hypothesis that $X$ causes $Y$.

	\subsection*{Proof of Proposition \ref{theorem_discrete}}
	We should prove equivalent results from Propositions \ref{prop:simconds}, \ref{prop:sim2id}, and \ref{thm:theone} for the discrete case. We define a discrete notion of similarity. 
	\begin{equation}
	    A \overset{d}{\sim} B \iff \exists \text{ bijective } g \text{ such that } g(A) \overset{I}{=} B.
	\end{equation}
	Note that Definition \ref{defn:id} should not be changed as it work in both discrete and continuous cases. We state our identifiability result in discrete case as follows:
	\begin{center}
	    \textit{Assumption \ref{assm:mechanismdiscrete} holds if and only if } $\forall a,b \in \mathcal{X}: \tilde{Y}_a \overset{d}{\sim} \tilde{Y}_{b}$.
	\end{center}
	The same reasoning in the proof of Proposition \ref{prop:simconds} yields this result.
	
	The result equivalent to Proposition \ref{prop:sim2id} is as follows:
	
	\begin{center}
	    \textit{For any two discrete-valued random variables $A$ and $B$, if $A \overset{d}{\sim} B$, then $\Psi_A \overset{I}{=} \Psi_B$.}
	\end{center}
	To prove this result, it suffices to change the proof of Proposition \ref{prop:sim2id} by replacing $|det(J_g(a))|$ with $1$ (in Equation \eqref{thatremark}), and use $p^i_V$ to denote the probability mass function (instead of probability density function). 
	
	Finally, we state the result equivalent to Proposition \ref{thm:theone}, for discrete case:
	\begin{center}
	    \textit{In the discrete case, for every $a,b \in \mathcal{X}$, we have $\Psi_{\tilde{Y}_a} \overset{I}{=} \Psi_{\tilde{Y}_b}$ if and only if $\Gamma_{X \to Y} \independent X$.}
	\end{center}
	The exact same reasoning in Proposition \ref{thm:theone} works for proving this result. Finally, Proposition \ref{theorem_discrete} is proved in the same way as Theorem \ref{theorem}.
	
	\subsection*{Proof of Proposition \ref{theorem_mult}}
\	Proposition \ref{theorem_mult} is a restatement of Theorem \ref{theorem} for multivariate case. By setting $X:= \mathbf{PA}_i$ and $Y := V_i$, the result is immediately implied from Theorem \ref{theorem}.
\end{document}